\documentclass[10pt]{article}
\usepackage{amsmath}
\usepackage{graphicx}
\usepackage[inline]{enumitem}
\usepackage{natbib}
\usepackage{url}
\usepackage{bbm}
\usepackage{bm}
\usepackage{amsfonts}
\usepackage{amsthm}
\usepackage{authblk}
\usepackage{hyperref}
\usepackage[verbose=true]{geometry}

\newgeometry{
    textheight=9in,
    textwidth=5.5in,
    top=1in,
    headheight=12pt,
    headsep=25pt,
    footskip=30pt
}

\usepackage{graphicx}
\usepackage{booktabs}
\usepackage{multirow}
\usepackage{array}
\usepackage{siunitx}
\usepackage{verbatim}
\usepackage{setspace}
\usepackage{etoolbox}
\usepackage{svg}
\usepackage{color, colortbl}
\usepackage{pgfplots}
\usetikzlibrary{calc}
\pgfplotsset{compat=1.18}
	
\definecolor{Gray}{gray}{0.8}

\graphicspath{{plots/}}
\newcommand{\E}{\mathbb{E}}
\newcommand{\Lagrangian}{\mathcal{L}}

\newcommand{\Var}{\mathrm{var}}

\DeclareMathOperator*{\argmin}{arg\,min}
\DeclareMathOperator*{\argmax}{arg\,max}

\newtheorem{theorem}{Theorem}

\newcommand{\mytitle}{Automatic debiasing of neural networks via moment-constrained learning}

\title{\mytitle}
\date{\today}
\author[1]{Christian L. Hines}
\author[2]{Oliver J. Hines}
\affil[1]{The Alan Turing Institute, London, UK}
\affil[2]{Columbia University, New York, NY, USA}

\begin{document}
\maketitle

\begin{abstract}

Causal and nonparametric estimands in economics and biostatistics can often be viewed as the mean of a linear functional applied to an unknown outcome regression function.  Naively learning the regression function and taking a sample mean of the target functional results in biased estimators, and a rich debiasing literature has developed where one additionally learns the so-called Riesz representer (RR) of the target estimand (targeted learning, double ML, automatic debiasing etc.). Learning the RR via its derived functional form can be challenging, e.g. due to extreme inverse probability weights or the need to learn conditional density functions. Such challenges have motivated recent advances in automatic debiasing (AD), where the RR is learned directly via minimization of a bespoke loss. We propose moment-constrained learning as a new RR learning approach that addresses some shortcomings in AD, constraining the predicted moments and improving the robustness of RR estimates to optimization hyperparamters. Though our approach is not tied to a particular class of learner, we illustrate it using neural networks, and evaluate on the problems of average treatment/derivative effect estimation using semi-synthetic data. Our numerical experiments show improved performance versus state of the art benchmarks.
\end{abstract}



\section{Introduction}

Several problems in causal inference, economics, and biostatistics can be viewed as inferring the average moment estimand $\Psi \equiv\E[m(\mu, W)]$, where $W = (Y,Z)$ consists of an outcome $Y$ and inputs $Z = (A,X)$ often comprised of a treatment $A$ and covariates $X$, with $\mu(z) \equiv \E[Y|Z=z]$, and $m(f,W)$ is a known functional that is linear in $f$. Examples of this setup include average treatment, policy and derivative effects, as outlined below. For such estimands, naively plugging in regression estimates $\hat{\mu}$ and taking the sample mean of $m(\hat{\mu}, W)$, given i.i.d. observations of $W$, generally leads to biased estimates which converge to $\Psi$ at less than the parametric $\sqrt{n}$ rate.

These biases arise because the bias-variance trade off of the regression estimator is controlled by a generic loss (e.g. mean squared error, cross-entropy) that does not adequately control for biases in the downstream estimation task. In particular, the true regression function $\mu$ satisfies $\E[\alpha(Z)\{Y-\mu(Z)\}] = 0$ for any function $\alpha$, but the same is not true of the empirical mean $\E_n[\alpha(Z)\{Y-\hat{\mu}(Z)\}]$, which may converge to zero slower than the $\sqrt{n}$ rate. Biases for average moment estimands take this form for an estimand-specific function $\alpha$. Specifically, the `plug-in bias' is characterized by the Riesz representer (RR) $\alpha$, which is an unknown function such that $\Psi = \langle \mu, \alpha \rangle$, where $\langle f, g \rangle \equiv \E[f(Z)g(Z)]$ denotes an inner product over a Hilbert space $\mathcal{H}$ equipped with norm $||f|| \equiv \langle f, f \rangle ^{1/2} < \infty$ and it is assumed that $\mu\in \mathcal{H}$. Existence of a unique $\alpha \in \mathcal{H}$ follows by Riesz's representation theorem since $f \mapsto h(f) \equiv \E[m(f, W)]$ is a bounded linear map. 

\textbf{Example 1: Average treatment effect (ATE)}.
\textit{
For a binary treatment $A \in \{0,1\}$, the ATE is the difference in mean outcome when an intervention assigns treatment versus no-treatment uniformly across the population \citep{Rosenbaum1983}. Under standard causal assumptions, the ATE is identified by $\Psi = \E[\mu(1,X) - \mu(0,X)]$, which is an average moment estimand for the moment functional $m(f, W) \equiv f(1, X) - f(0,X)$. Letting $p(x) \equiv \E[A|X=x]$, and assuming $p(x) \in (0,1)$, the ATE has the RR $\alpha(z) = \{a-p(x)\} / [ p(x) \{1-p(x)\}]$.
}

\textbf{Example 2: Average policy effect (APE)}
\textit{
Using the setup from Example 1, the APE considers the mean outcome when an intervention assigns treatment according to a known treatment policy $x\mapsto \pi(x) \in \{0,1\}$ \citep{Dudik2011,VanderLaan2014,Athey2021}. The APE is identified by $\Psi = \E[\pi(X) \{\mu(1,X) - \mu(0,X)\} + \mu(0,X)]$, which is an average moment estimand for the moment functional $m(f, W) \equiv \pi(X) \{f(1,X) - f(0,X)\} + f(0,X)$. The APE has the RR $\alpha(z) = [\pi(x)\{a-p(x)\} + p(x)\{1-a\}]/[ p(x) \{1-p(x)\}]$.
}

\textbf{Example 3: Average derivative effect (ADE)}
\textit{
For a continuous treatment $A \in \mathbb{R}$, the ADE, $\Psi = \E[\mu^\prime(A,X)]$ is average derivative of the conditional response function, where superscript prime denotes the derivative w.r.t. $a$, and we assume that $\mu^\prime$ exists \citep{Hardle1989,Newey1993,Imbens2009,Rothenhausler2019}. The ADE is an average moment estimand for the moment functional $m(f, W) \equiv f^\prime(A, X)$. Letting $p(a|x)$ denote the conditional density of $A$ given $X$, and assuming $p(a|x) > 0$, and $p(a|x) = 0$ for $a$ on the boundary of the support of $A$, then the ADE has the RR $\alpha(z) = -p^\prime(a|x) / p(a|x)$.
}

\textbf{Example 4: Incremental policy effect (IPE)}
\textit{
Using the setup from Example 3, the IPE \citep{Athey2021} is $\Psi = \E[\pi(X)\mu^\prime(A,X)]$, where $x\mapsto \pi(x) \in [-1,1]$ is a known policy function. The IPE is an average moment estimand for the moment functional $m(f, W) \equiv \pi(X)f^\prime(A, X)$ and has the RR $\alpha(z) = -\pi(x)p^\prime(a|x) / p(a|x)$.
}

Following semiparametric efficiency results \citep{Robins1994,Newey1994}, a rich literature has developed in recent years that compensates for plug-in biases either by estimating the RR then shifting the naive estimator (double machine learning \citep{Chernozhukov2018}), or retrospectively modifying the estimates $\hat{\mu}$ such that the estimated plug-in bias is negligible (targeted learning \citep{van_der_laan_cross-validated_2011}). Both approaches are celebrated for constructing efficient estimators that converge at $\sqrt{n}$ rate even when learners for the conditional mean outcome and the RR converge at a slower e.g. $n^{1/4}$ rate. 

As exemplified above, the RR can be a complicated function of the data distribution making learning the RR using its derived form challenging. For instance, the RR of the ATE and APE can be estimated using a learner $\hat{p}$ of the propensity score $p$, but the resulting estimates may be overly sensitive to the error $\hat{p}(x) - p(x)$ when $\hat{p}(x)$ is close to 0 or 1, since $p$ appears in the denominator of the RR. Similarly, RR estimators for the ADE typically use kernel estimators and are overly sensitive to the bandwidth \citep{Cattaneo2013}.

To overcome such issues, recent work has sought to learn the RR directly from the data, without using knowledge of its functional form. Initial approaches for binary treatments used balancing weights rather than propensity scores to estimate plug-in biases \citep{Zubizarreta2015,Athey2018}. These approaches have been generalized through the adversarial RR learner of \citep{Chernozhukov2020} that builds on the (also adversarial) augmented minimax linear estimator \citep{Hirshberg2021} and similar estimators for conditional moment models \cite{Dikkala2020}. More recently, automatic debiasing (AD) \citep{Chernozhukov2021,Chernozhukov2022} has been proposed to bypass the need to solve a computationally challenging adversarial learning problem by constructing a simple loss that is equivalent to minimizing the mean squared error in the RR. AD generalizes similar approaches using approximately sparse linear regression \cite{Chernozhukov2022b} and reproducing kernel Hilbert spaces \cite{Singh2023}.

Despite the success of AD, there are several areas for improvement which we address in our work. First, the AD loss is unbounded, and includes a negative average moment term that can lead to extreme moment predictions in the final RR estimator. In practice, early stopping using an external validation set is recommend to avoid such issues, however the resulting learners may be overly sensitive to e.g. early stopping and learning rate hyperparameters. Second, oftentimes the RR admits known inner products which are ignored by the AD loss. For instance, we know \emph{a priori} that for the ADE/ATE $h(a) = \E[A\alpha(Z)] = 1$. Methods which estimate the RR using its derived form approximately encode such identities, but this is not the case when the RR is learned by AD.


\textbf{Contributions:} We propose average moment estimators based on a new decomposition of the RR in terms of the moment-constrained function $\beta_\perp(z)$. Specifically, $\beta_\perp$ minimizes the mean squared error in predicting a known function $\beta(z)$ subject to $h(\beta_\perp) = 0$, where $\beta$ is chosen such that one knows \emph{a priori} that $h(\beta) \neq 0$. We propose an approach to learning $\beta_\perp$ and derive debiased estimators of $\Psi$ based on initial ML estimates $\hat{\mu}$ and $\hat{\beta}_\perp$.

The advantage of learning the RR via $\beta_\perp$ rather than the AD loss is that the resulting RR estimates better control for extreme out-of-sample RR predictions since constants of proportionality in the RR are estimated using the estimation sample rather than the training sample. Moreover, our proposed estimator for $\beta_\perp$ is robust to overfitting issues that may arise when using the AD loss and thus is less sensitive to the tuning of optimization/model hyperparameters. Our proposal remains `automatic' in the sense of not requiring the functional form of the RR to be derived. However, unlike AD, which only requires knowledge of the moment function $m$, we additionally require a known function $\beta$ with $h(\beta) \neq 0$. The need to construct such a function may be viewed as a limitation of our proposal, though we contend that doing so is straightforward, as we demonstrate for Examples 1 to 4.

Though our approach is not tied to a particular machine learning method, we evaluate our estimators on two semi-synthetic datasets using multi-tasking neural networks, making comparisons with RieszNet \citep{Chernozhukov2022} for ADE/ATE estimation, and DragonNet \citep{Shi2019}, Reproducing Kernel Hilbert Space (RKHS) Embedding \citep{Singh2023}, Neural Net (NN) Embedding \citep{Xu2022} for ATE estimation. 
To ensure a fair comparison we re-implement RieszNet and DragonNet learners and estimators, with reproduction code available at \url{https://github.com/crimbs/madnet}.

\section{Estimation}

\subsection{Debiased estimation}
\label{sect:standard_theory}

Given a sample of $n$ i.i.d. observations, estimators of $\Psi$ are typically based on the empirical distribution $\mathbb{E}_n[.] = n^{-1}\sum_{i=1}^n (.)_i$, the regression function $\hat{\mu}$, and the estimated RR function $\hat{\alpha}$. An estimator $\hat{\psi}$ of $\Psi$ is said to be regular asymptotically linear (RAL) if its error behaves like an empirical process, i.e. $\sqrt{n}(\hat{\psi} - \Psi)= G_n[\varphi(W)] + o_p(1)$ for some finite variance function $\varphi(W)$, where $G_n[.] \equiv \sqrt{n} (\mathbb{E}_n[.]- \E[.])$ is the empirical process operator. RAL estimators are unbiased since, by the central limit theorem, $\sqrt{n}(\hat{\psi} - \Psi)$ converges (in distribution) to a mean zero normal random variable with variance $\Var[\varphi(W)]$. Results from nonparametric efficiency theory \citep{Newey1994} imply that $\Var[\varphi(W)]$ is minimized when $\varphi(W) = m(\mu, W) + \alpha(Z)\{Y - \mu(Z)\}$ is the uncentered influence curve \citep{Hampel1974,Ichimura2022} of $\Psi$, also called the pseudo-outcome \citep{Kennedy2020, Hines2023b} (see \cite{Hines2022, Kennedy2022} for pedagogical reviews). Thus, one can construct standard errors for $\hat{\psi}$ by approximating $\varphi$ with some $\hat{\varphi}$ and taking a sample variance. To consider specific estimators we use the identity
\begin{align}
    \sqrt{n}(\hat{\psi} - \Psi) = G_n \left[\varphi(W) \right] - \underbrace{\sqrt{n}\E_n \left[\hat{\varphi}(W) - \hat{\psi} \right]}_{\text{plug-in bias}} + \underbrace{\sqrt{n}\E[\hat{\varphi}(W) - \Psi]}_{\text{first-order remainder}} +  \underbrace{ G_n \left[\hat{\varphi}(W) - \varphi(W) \right] }_{\text{second-order remainder}} \label{von_miesz}
\end{align}
which can be shown to hold (by canceling terms on the right hand side) for any $\hat{\psi}$ and pair of measurable functions $\varphi, \hat{\varphi}$. The second-order remainder above is usually not a concern and is $o_p(1)$ under weak assumptions, e.g. when $\E[\{\varphi(W) - \hat{\varphi}(W)\}^2] = o_p(1)$ and $\hat{\varphi}$ is obtained from an independent sample. In practice, the latter assumption motivates estimators which apply some form of sample-splitting/cross-fitting to estimate $\hat{\varphi}$ and evaluate the estimator \citep{van_der_laan_cross-validated_2011,Chernozhukov2018}.

Letting $h_n(f) \equiv \mathbb{E}_n [m(f, W)]$, a naive estimator is $\hat{\Psi}^{(\text{Direct})} \equiv h_n(\hat{\mu})$, which does not use the RR estimates $\hat{\alpha}$. To examine the bias properties of the naive estimator, consider \eqref{von_miesz} when $\hat{\psi} = \hat{\Psi}^{(\text{Direct})}$ and $\hat{\varphi}(W) = m(\hat{\mu}, W) + \hat{\alpha}(Z)\{Y - \hat{\mu}(Z)\}$. Following \citep{Chernozhukov2020}, the first-order remainder reduces to the `mixed bias', $- \sqrt{n} \langle \hat{\mu} - \mu, \hat{\alpha} - \alpha  \rangle$, the square of which is bound by Cauchy-Schwarz as $n  \langle \hat{\mu} - \mu, \hat{\alpha} - \alpha  \rangle^2 \leq n||\hat{\mu} - \mu||^2 ||\hat{\alpha} - \alpha||^2$. The first-order remainder will therefore be $o_p(1)$ when $\hat{\mu}$ or $\hat{\alpha}$ converge to their true counterparts at sufficiently fast rates with sample size. Moreover, one can trade off accuracy in $\hat{\mu}$ and $\hat{\alpha}$, a property known as rate double robustness.

The plug-in bias $\sqrt{n}\mathbb{E}_n [\hat{\varphi}(W) - \hat{\Psi}^{(\text{Direct})} ] = \sqrt{n}\mathbb{E}_n[\hat{\alpha}(Z)\{Y - \hat{\mu}(Z)\}]$, however, is generally not $o_p(1)$, and hence $\hat{\Psi}^{(\text{Direct})}$ is not RAL. The main challenge in obtaining RAL estimators is therefore removing plug-in biases and there are two main strategies for doing so. One-step debiased estimators simply add the plug-in bias to both sides of \eqref{von_miesz}, resulting in the (double robust) RAL estimator $\hat{\Psi}^{(\text{DR})} \equiv \mathbb{E}_n[\hat{\varphi}(W)]$, or equivalently
\begin{align}
    \hat{\Psi}^{(\text{DR})} &= \underbrace{h_n(\hat{\mu})}_{\text{direct estimator}} + \underbrace{\mathbb{E}_n[\hat{\alpha}(Z)\{Y - \hat{\mu}(Z)\}]}_{\text{bias correction}}. \label{eq_dr}
\end{align}

Targeted maximum likelihood estimators (TMLEs) 
are direct estimators of the form $\hat{\Psi}^{(\text{TMLE})} \equiv h_n(\hat{\mu}^*)$ where $\hat{\mu}$ is replaced with a `targeted' alternative $\hat{\mu}^*$ that solves 
\begin{align}
    \mathbb{E}_n[\hat{\alpha}(Z)\{Y - \hat{\mu}^*(Z)\}] = 0. \label{tmle_score}
\end{align}
Like one-step debiased estimators, TMLEs are double robust, with the mixed bias condition $\langle \hat{\mu}^* - \mu, \hat{\alpha} - \alpha  \rangle = o_p(n^{-1/2})$. Targeting can be achieved in many ways, for example by first defining the linear parametric submodel $\hat{\mu}_t(z) = \hat{\mu}(z) + t \hat{\alpha}(z)$ for a univariate indexing parameter $t\in \mathbb{R}$. Then obtaining an optimal $t^* = \argmin_{t \in \mathbb{R}} \mathbb{E}_n[\{Y-\hat{\mu}_t(Z)\}^2]$ which improves the fit of the outcome learner in the estimation sample, and ensures that $\hat{\mu}^* = \hat{\mu}_{t^*}$ is a solution to \eqref{tmle_score}. For the linear parametric submodel $t^* = \mathbb{E}_n[\hat{\alpha}(Z)\{Y - \hat{\mu}(Z)\}] / \mathbb{E}_n [ \hat{\alpha}^2(Z)]$ and $\hat{\Psi}^{(\text{TMLE})} = h_n(\hat{\mu}) + t^* h_n(\hat{\alpha})$, or equivalently
\begin{align}
    \hat{\Psi}^{(\text{TMLE})} = \underbrace{h_n(\hat{\mu})}_{\text{direct estimator}} + \underbrace{ \left(\frac{h_n(\hat{\alpha})}{\mathbb{E}_n [ \hat{\alpha}^2(Z)]}\right)}_{\text{scale factor}}  \underbrace{\mathbb{E}_n[\hat{\alpha}(Z)\{Y - \hat{\mu}(Z)\}]}_{\text{bias correction}} . \label{tmle}
\end{align}
Comparing $\hat{\Psi}^{(\text{TMLE})}$ with $\hat{\Psi}^{(\text{DR})}$ in \eqref{eq_dr}, we see that the TMLE introduces a scale factor $h_n(\hat{\alpha}) / \mathbb{E}_n [ \hat{\alpha}^2(Z)]$, which is an empirical approximation to the population value $h(\alpha) / ||\alpha||^2 = 1$, thereby rescaling the bias correction of the DR estimator. Variations of the TMLE method often include canonical generalized linear model (GLM) link functions in the parametric submodel definition, and maximize the associated GLM log-likelihood (hence the name TMLE), see e.g. \citep{VanDerLaan2016} for submodel proposals. GLM variations of this type may be used e.g. when $Y$ is binary and a cross-entropy outcome loss is preferred. Finally, to motivate new RR learning methodologies, we remark that \eqref{tmle_score} and \eqref{tmle} are invariant to constants of proportionality in $\hat{\alpha}$, thus one might consider RR learners that are agnostic to such constants.



\subsection{Debiased estimation with moment constraints}

\begin{figure}[htbp]
    \centering
    \includegraphics[width=0.35\textwidth]{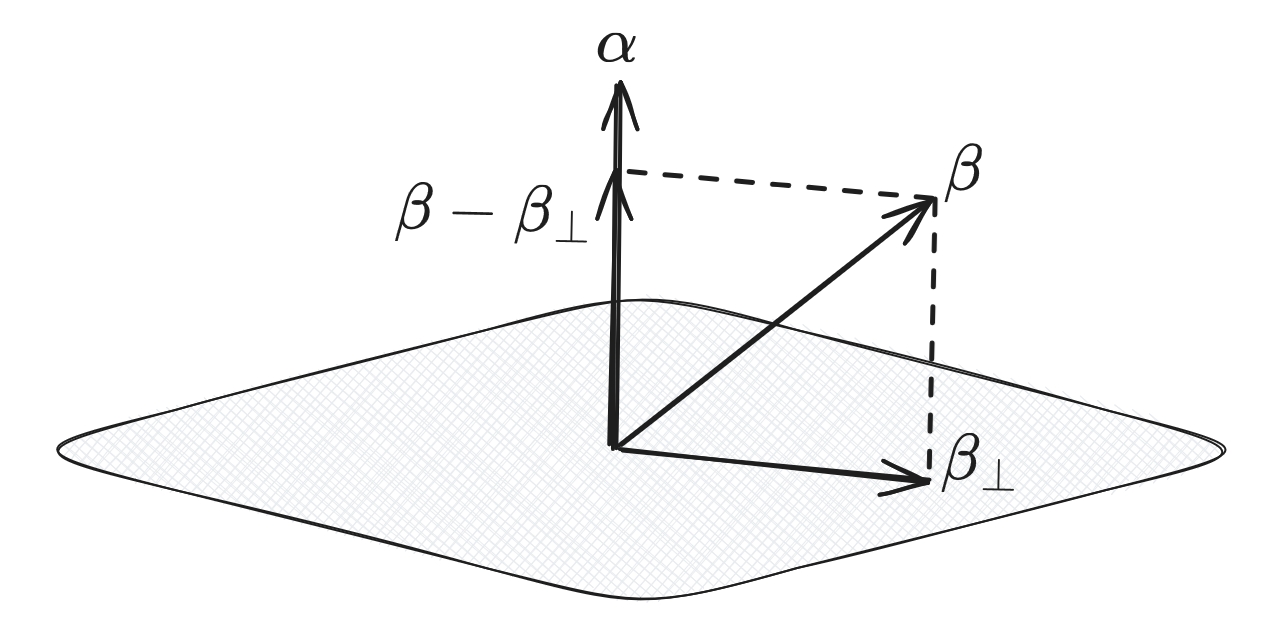}
    \caption{Illustration of moment-constrained functions. The plane represents the space of zero average moment functions, i.e. $f$ such that $h(f) = \langle f, \alpha \rangle = 0$. The non-zero function $\beta - \beta_\perp$ is orthogonal to the plane, and thus is a scalar multiple of $\alpha$.
    }
    \label{fig:geometrical_alt}
\end{figure}

Our main contribution is to propose average moment estimators based on the identity
\begin{align}
    \alpha(z) &= \frac{h(\beta)}{||\beta-\beta_\perp||^2}\{\beta(z) - \beta_\perp(z)\} \label{new_ident_1}
\end{align}
where $\beta \in \mathcal{H}$ is a known function with $h(\beta) \neq 0$ and $\perp$ denotes projection on to orthogonal complement set $\mathcal{C}^\perp \equiv \{f\in \mathcal{H} \mid  h(f) = 0 \}$. Specifically, $\beta_\perp = \argmin_{f \in \mathcal{C}^\perp} ||\beta - f||$. A geometric illustration of this result is provided in Figure \ref{fig:geometrical_alt}.

\textbf{Proof of \eqref{new_ident_1}}:
\textit{
Note that $\mathcal{C}^\perp = \{f\in \mathcal{H} \mid  \langle f, \alpha \rangle = 0 \}$. By Hilbert's projection theorem, $\beta_\perp$ exists, with
\begin{align*}
    \beta_\perp(z) \equiv \beta(z) - \frac{\langle \beta, \alpha \rangle}{||\alpha||^2} \alpha(z) \quad\iff\quad \alpha(z) = \frac{||\alpha||^2}{h(\beta)}\{\beta(z) - \beta_\perp(z)\}
\end{align*}
where we use $\langle \beta, \alpha \rangle = h(\beta)$. Taking the norm of both sides and solving for $||\alpha|| \neq 0$ gives $||\alpha|| = |h(\beta)|/||\beta-\beta_\perp||$ which completes the proof.
}

The identity in \eqref{new_ident_1} offers new avenues for debiased estimation of $\Psi$ via learning $\beta_\perp$ and $\mu$. One limitation of this proposal, however, is that we require construction of a function $\beta \in \mathcal{H}$ such that $h(\beta) \neq 0$ \emph{a priori} (and $||\beta|| < \infty$ since $\beta \in \mathcal{H}$). Due to the variety of possible moment functionals $m(f, W)$, we do not offer a general algorithm for constructing such functions, however, we have found in practice that there are usually candidates for $\beta$, where $m(\beta, W)$ is a known constant, and hence $h(\beta)$ is known \emph{a priori}. For instance, for the ADE we set $m(\beta, W) = \beta^\prime(A, X) = 1$ and integrate to obtain $\beta(z) = \beta(a, x) = a$ with $h(\beta) = 1$ \emph{a priori}. 
This choice is not unique, however, since $\beta(z) = \exp(a)$ is also a valid choice for the ADE, with $h(\beta) = \E[\exp(A)] \neq 0$. As for the other estimands in Examples 1 to 4: for the ATE $h(\beta) = 1$ when $\beta(z) = a$; for the APE, $h(\beta) = 1$ when $\beta(z) = a + 1 - \pi(x)$; for the IPE, $h(\beta) = 1$ when $\beta(z) = a /\pi(x)$, or if there is concern that $\pi(X)$ can be zero, one can instead let $\beta(z) = a /\pi(x) $ when $\pi(x) \neq 0$ and $\beta(z) = 0$ otherwise, then estimate $h(\beta) = \Pr[\pi(X) \neq 0]$. For full generality we develop estimators in the setting where $h(\beta)$ must be estimated, but our results simplify slightly when $h(\beta)$ is known.

\textbf{Example ATE:} \textit{
Denoting $\beta(z) = \beta(a,x) = a$, the known RR and \eqref{new_ident_1} imply
\begin{align*}
    \beta_\perp(a,x) = p(x) + \{a - p(x)\}\left(1 - \frac{1}{p(x)\{1 - p(x)\}}\E\left[\frac{1}{p(X)\{1 - p(X)\}}\right]^{-1}\right). 
\end{align*}
It is insightful to compare $\beta_\perp$, which minimizes $\E[\{A - f(A,X)\}^2]$ given $\E[f(1,X)] = \E[f(0,X)]$, with $p(x)$, which minimizes the same mean squared error, under the stronger constraint $f(1,X) = f(0,X)$. We notice that $p(x)$ lies on the interval $(0,1)$, but the same is not true of $\beta_\perp$, which has weaker restrictions on its outputs: $\beta_\perp(1,x) < 1$ and $\beta_\perp(0,x) > 0$. Also $p$ and $\beta_\perp$ are related by the identity $p(x) = \E[\beta_\perp(A,X)|X=x]$. 
}

\textbf{Example ADE:} \textit{
Denoting $\beta(z) = \beta(a,x) = a$, the known RR and \eqref{new_ident_1} imply
\begin{align*}
    \beta_\perp(a,x) = a - \E\left[\left(\frac{p^\prime(A|X)}{p(A|X)} \right)^2 \right]^{-1} \frac{p^\prime(a|x)}{p(a|x)},
\end{align*}
which minimizes $\E[\{A - f(A,X)\}^2]$ given $\E[f^\prime(A, X)] = 0$. As in the previous example, we remark that $\E[\beta_\perp(A,X)|X=x] = \E[A|X=x]$.
}

In Section \ref{sect:moment_constrained_learning} below, we propose methods for learning $\beta_\perp$, however, we first show how the standard debiased estimators from Section \ref{sect:standard_theory} look when debiasing is achieved using initial estimates $\hat{\beta}_\perp$ instead of $\hat{\alpha}$. Specifically, we consider
\begin{align*}
    \hat{\alpha}(z) &= \frac{h_n(\beta-\hat{\beta}_\perp) \{\beta(z) - \hat{\beta}_\perp(z)\} }{ \mathbb{E}_n [\{\beta(Z) - \hat{\beta}_\perp(Z)\}^2] }. 
\end{align*}
Under this parameterization, the DR
estimator in \eqref{eq_dr} and TMLE in \eqref{tmle} both become
\begin{align}
\hat{\Psi}^{(\perp, \text{DR})} &\equiv \underbrace{h_n(\hat{\mu})}_{\text{direct estimator}} + \underbrace{ \left(\frac{h_n(\beta-\hat{\beta}_\perp)}{\mathbb{E}_n [\{\beta(Z) - \hat{\beta}_\perp(Z)\}^2]} \right)}_{\text{scale factor}} \underbrace{ \mathbb{E}_n [\{\beta(Z) - \hat{\beta}_\perp(Z)\}\{Y - \hat{\mu}(Z)\}] }_{\text{unscaled bias correction}},
\end{align}
which is of the same form as \eqref{tmle} but with $\beta - \hat{\beta}_\perp$ replacing $\hat{\alpha}$. Moreover, since the the TMLE score equation in \eqref{tmle_score} only requires estimating the RR up to constants of proportionality, TMLEs can be derived using parametric submodels where $\hat{\alpha}$ is replaced with $\beta - \hat{\beta}_\perp$, e.g. by using the linear submodel $\hat{\mu}_t(z) = \hat{\mu}(z) + t \{\beta(z) - \hat{\beta}_\perp(z)\}$, which gives $h_n(\hat{\mu}^*) = \hat{\Psi}^{(\perp, \text{DR})}$. Theorem \ref{main_theorem} gives general conditions under which a TMLE $h_n(\hat{\mu}^*)$ based on $\hat{\beta}_\perp$ is RAL, with $\hat{\Psi}^{(\perp, \text{DR})}$ being a special case. Essentially, this theorem controls the first-order remainder of \eqref{von_miesz} using the mixed-bias condition $\langle \hat{\mu}^* - \mu , \hat{\beta}_{\perp}  - \beta_\perp\rangle = o_p(n^{-1/2})$, and uses empirical process assumptions to control the second-order remainder of \eqref{von_miesz}.





\begin{theorem}
    \label{main_theorem}
    Let $\hat{\beta}_\perp$ be an estimator for $\beta_\perp$, and let $\hat{\mu}^*$ be an estimator for $\mu$ that is targeted such that $\E_n[\{\beta(Z) - \hat{\beta}_\perp(Z)\}\{Y - \hat{\mu}^*(Z)\}] = 0$. Assume that each of the following terms are $o_p(1)$: $\sqrt{n}\langle \hat{\mu}^* - \mu , \hat{\beta}_{\perp}  - \beta_\perp\rangle$, $G_n[m(\hat{\mu}^* - \mu, W)]$, $G_n\left[\{\hat{\beta}_\perp(Z) - \beta_\perp(Z)\}\{\hat{\mu}^*(Z) - \mu(Z)\}\right]$, $G_n\left[\{\beta(Z) - \beta_\perp(Z)\}\{\hat{\mu}^*(Z) - \mu(Z)\}\right]$, and $G_n\left[\{\hat{\beta}_\perp(Z) - \beta_\perp(Z)\}\{Y - \mu(Z)\}\right]$.
    Then $h_n(\hat{\mu}^*)$ is a RAL estimator of $\Psi$ with uncentered influence curve $\varphi(W)$, and hence $\sqrt{n}(h_n(\hat{\mu}^*)-\Psi)$ converges in distribution to a mean-zero normal random variable with variance $\Var[\varphi(W)]$. Proof in Supplement \ref{main_theorem_proof}.
\end{theorem}

\subsection{Moment-constrained learning}
\label{sect:moment_constrained_learning}

We propose learners for the moment-constrained function $\beta_\perp$ using the property that $\beta_\perp$ is the function $f\in\mathcal{H}$ that solves
\begin{align}
    \text{minimize:}& \quad ||\beta - f||^2\nonumber \\
    \text{subject to:}& \quad h(f) = 0 \label{constraint}
\end{align}
Similar constrained learning problems have been studied in the context of ML with fairness constraints \citep{Nabi2024}. E.g. \cite{Zafar2017, Akhtar2021} minimize a classification loss, while ensuring that predictions are uncorrelated with specific sensitive attributes (race, sex etc.). The primal problem in \eqref{constraint} is characterized by the Lagrangian
\begin{align}
    \Lagrangian(f, \lambda) \equiv \E\left[\{\beta(Z) - f(Z)\}^2 + \lambda m(f, W)\right] \label{pop_lagrangian}
\end{align}
where $\lambda \in \mathbb{R}$ is a Lagrange multiplier, 
and a solution is obtained by finding $f^*$ and $\lambda^*$ such that $\Lagrangian(f^*, \lambda^*) =\max_{\lambda \in {\mathbb{R}}} \min_{f \in \mathcal{H}} \Lagrangian(f, \lambda)$. Naively, therefore, one might learn $f$ by performing gradient descent over parameters indexing $f$ and gradient ascent on $\lambda$, as in the basic differential multiplier method (BDMM) of \cite{Platt1987}.

In our numerical experiments, we consider the setting where $f = f_w$ is the output of a multilayer perceptron (MLP) with weights $w$. We observe that application of BDMM to a sample analogue of $\Lagrangian(f_w, \lambda)$ leads to empirical constraint violations that oscillate around zero, as the number of ascent/descent iterations increases (shown in Figure~\ref{fig:damping-plot} of the supplement). Similar behavior is documented elsewhere for adversarial function learners \citep{Schafer2019, Mokhtari2020}. Instead, stable constraint violations were achieved by effectively replacing the constraint in \eqref{constraint} with the equivalent constraint $|h(f)| \leq 0$, yielding the Lagrangian
\begin{align}
    \Tilde{\Lagrangian}(f, \Tilde{\lambda}) \equiv \E\left[\{\beta(Z) - f(Z)\}^2\right] + \Tilde{\lambda} |h(f)| \label{pop_star_lagrangian}
\end{align}
with empirical MLP analogue $\Tilde{\Lagrangian}_n(f_w, \Tilde{\lambda}) \equiv \E_n\left[\{\beta(Z) - f_w(Z)\}^2\right] + \Tilde{\lambda} |h_n(f_w)|$.
In this formulation, $\Tilde{\lambda} \geq 0$ penalizes the sample average moment of $f_w$ in a similar way to the smoothing parameters in conventional Lasso/ridge regression. The key difference between these classical methods, however, is that the penalty $|h_n(f_w)|$ depends on the observed data, and not only on the weights $w$. In practice we set $\Tilde{\lambda}$ to a constant value during training, and minimize over $w$ using gradient descent, though one might consider alternative methods e.g. where $\Tilde{\lambda}$ increases monotonically with the number of descent iterations (epochs).

\subsection{Comparison with Automatic debiasing (AD)}

AD \citep{Chernozhukov2021, Chernozhukov2022} is an RR learning method based on the identity
\begin{align*}
    \alpha &= \argmin_{\hat{\alpha}\in \mathcal{H}} || \alpha - \hat{\alpha}||^2 \\
    &= \argmin_{\hat{\alpha}\in \mathcal{H}} || \hat{\alpha}||^2 -2 \langle \hat{\alpha}, \alpha \rangle \\
    &= \argmin_{\hat{\alpha}\in \mathcal{H}} \E\left[\hat{\alpha}(Z)^2 - 2m(\hat{\alpha},W) \right].
\end{align*}
The AD RR learner minimizes a sample analogue of this expectation.
We connect AD to our proposal as follows. Consider that $\hat{\alpha}\in\mathcal{H}$ can be written as $\hat{\alpha}(z) = 2\lambda^{-1}\{\beta(z) - f(z)\}$ where $\lambda \neq 0$ is constant and $f\in \mathcal{H}$. Thus, the AD population minimization becomes
\begin{align*}
    \beta_\lambda &\equiv \argmin_{f \in \mathcal{H}} \E\left[4\lambda^{-2}\{\beta(Z) - f(Z)\}^2 - 4\lambda^{-1} m(\beta - f,W) \right]\\
    &=\argmin_{f \in \mathcal{H}} \E\left[\{\beta(Z) - f(Z)\}^2 + \lambda m(f,W) \right].
\end{align*}
with $\alpha(z) = 2\lambda^{-1}\{\beta(z) - \beta_\lambda(z)\}$. The objective above is the Lagrangian $\Lagrangian(f, \lambda)$ in \eqref{pop_lagrangian}. In this construction $\lambda$ is unrestricted, therefore provided that $\lambda_\perp \equiv \argmax_{\lambda \in \mathbb{R}}\min_{f \in \mathcal{H}} \Lagrangian(f, \lambda)$ is finite and non-zero, one can write $\alpha(z) = 2\lambda_\perp^{-1}\{\beta(z) - \beta_{\lambda_\perp}(z)\}$. Appealing to the primal problem in \eqref{constraint}, we see that $\beta_{\lambda_\perp} = \beta_\perp$ and, $\lambda_\perp = 2 ||\beta - \beta_\perp||^2/h(\beta)$ as in \eqref{new_ident_1}.

Moment-constrained learning, therefore, reinterprets the AD loss as a Lagrangian when one has access to a function $\beta$ such that $h(\beta) \neq 0$, and one is agnostic to constants of proportionality in the RR. When estimating $\Psi$, these constants are estimated using the estimation sample rather than by the RR learner directly. By connecting the RR to the primal problem, one is able to leverage constrained learning methods that may have better empirical performance, e.g. using the Lagrangian $\Tilde{\Lagrangian}$ in \eqref{pop_star_lagrangian}. Moreover, this connection hints at future theoretical study of RRs and AD via constrained function learning theory.



\section{MADNet: Moment-constrained Automatic Debiasing Networks}

\begin{figure}[htbp]
    \centering
    \includegraphics[width=0.75\textwidth]{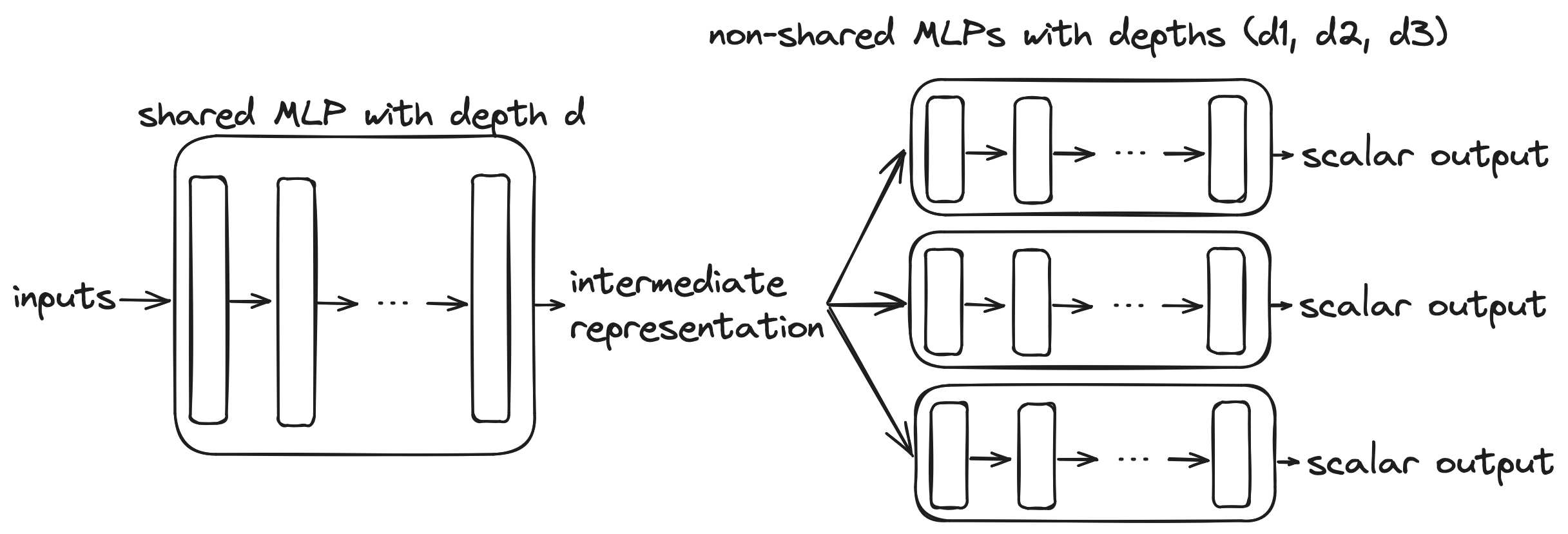}
    \caption{Multi-headed MLP architecture with three outputs. Typically the intermediate representation has the same width as the internal layers of the shared MLP, and the non-shared MLPs have internal layers with half the width of the shared MLP. During training, a single loss is used based on all scalar outputs, and MLP weights are learned using back-propagation over the entire multi-headed MLP.
    }
    \label{fig:sharedmlps}
\end{figure}

Multi-headed MLPs, as illustrated through the schematic in Figure \ref{fig:sharedmlps}, are emerging as a popular architecture for estimating average moment estimands using deep learning. Initial efforts focused on the binary treatment setting, such as the multi-headed MLP outcome learner TARNet (Treatment Agnostic Representation Network) \citep{Shalit2017}. TARNet takes inputs $X$ and produces two scalar outputs representing $\mu(1, X)$ and $\mu(0, X)$ respectively. During training, an outcome prediction error (e.g. mean-squared error) is minimized, with predictions of $\mu(A,X)$ obtained from one of the scalar outputs according to whether an observation is treated/untreated. The resulting outcome learner can be used to obtain plug-in estimates for e.g. the ATE/APE, optionally with debiasing by a separate RR learner as described in Section \ref{sect:standard_theory}.

DragonNet \citep{Shi2019} also focuses on binary treatments, extending TARNet by introducing a third scalar output MLP of zero depth, which is used to estimate the propensity score $p(X)$, and hence the RR of the ATE/APE. The authors reason that the propensity score MLP should have zero depth so that the shared MLP learns representations of $X$ that are predictive of the RR, since, for ATE estimation it is sufficient to learn the outcome conditional on $A$ and $p(X)$ only. This approach is generalized by \citet[Lemma 3.1]{Chernozhukov2022}, where it is shown that: for estimation of $\Psi$ it is sufficient to learn the outcome conditional on the RR only, i.e $E[Y|\alpha(Z) = \alpha(z)]$. Similarly, for estimation of $\Psi$, we show (Supplement \ref{proof_of_suff}) that it is sufficient to learn the outcome conditional on $\beta(Z) - \beta_\perp(Z)$ only. Moreover, we show (Supplement \ref{proof_of_ortho}) that $\mu(Z) = \mu_\perp(Z) + \Psi \{\beta(Z) - \beta_\perp(Z)\} / h(\beta)$
where $\mu_\perp$ is the projection of $\mu$ on $\mathcal{C}^{\perp}$. This result further highlights the role of the RR when learning the outcome for average moment estimation.

RieszNet is similar in structure to DragonNet, except with input $Z = (A, X)$ rather than $X$. Both also use a multi-tasking loss to learn $\mu$ and $\alpha$ simultaneously. We propose MADNet which uses the same network structure as RieszNet and a multi-tasking loss to learn $\mu$ and $\beta_\perp$ simultaneously. Specifically we consider the loss $\Tilde{\Lagrangian}_n(f_{w, 1}, \Tilde{\lambda}) + \rho\,\text{REGLoss}_n(f_{w, 2})$
where $\rho \geq 0$ is a hyperparameter, $\text{REGLoss}_n$ is a regression loss, e.g. the mean-squared error in the outcome prediction $\text{REGLoss}_n(f) = \E_n[\{Y - f(Z)\}^2]$, and $f_w(z) = (f_{w,1}(z), f_{w,2}(z))$ represents two outputs from a multi-headed MLP. Note that e.g. $f_{w,1}$ depends only on the weights of the shared MLP and the first non-shared MLP, but we write it as a function of all multi-headed MLP weights $w$ for convenience. Like RieszNet, for ATE/ADE estimation we replace $f_{w, 2}(z)$ in the regression loss with $\tilde{a} f_{w, 2}(z) + (1-\tilde{a})f_{w, 3}(z)$, where $\tilde{a}$ represents the min-max normalized treatment $a$ scaled on to the interval $[0, 1]$, i.e. $\tilde{a}_i = \{a_i - \min_n(a_i)\} / \{\max_n(a_i) - \min_n(a_i)\}$.



\textbf{Convergence rates}: The standard theory in Section \ref{sect:standard_theory} controls first-order remainders by requiring estimators to converge to their true counterparts at sufficiently fast rates. For neural network learners, recent convergence rate results have been obtained using the theory of critical radii \citep{Wainwright2019,Foster2023,Chernozhukov2020}. In particular, results are provided for MLPs with Rectified Linear Unit (ReLU) activation functions \citep{Farrell2021}, describing $L_2$ convergence rates in terms of the number of training observations and the network width/depth. Similar results exist for AD learners with moments satisfying the mean-squared continuity property $\E[\{m(u, W) - m(v, W)\}^2] \leq M ||u - v||^2$ for $M\geq 1$ and $u, v \in \mathcal{A}_n$, where $\mathcal{A}_n$ is a function set described by \citet{Chernozhukov2021}. We expect similar theoretical guarantees to hold in the current context due to the connection of AD with moment-constrained learning discussed above.


\textbf{Sample-splitting:} To obtain valid inference from a single sample, the standard theory in Section \ref{sect:standard_theory} relies on cross-fitting to control the second-order remainder terms in \eqref{von_miesz}. Cross-fitting is used to bypass Donsker class assumptions \citep{Bickel1982,van_der_laan_cross-validated_2011,Chernozhukov2018}, which restrict the complexity of the initial estimators and are usually not satisfied by ML algorithms. Recent work has sought to bypass Donsker conditions by instead relying on entropic arguments \citep{VanDeGeer2014} or leave-one-out stability \citep{Chen2022}. In practice DragonNet and RieszNet do not use sample splitting due to the associated computational burden. Instead, implementations of both algorithms split the data into training and validation sets, with the validation set used to control early stopping of the training algorithm. For estimation, the full dataset is used (training + validation). We also use this strategy for moment-constrained learning.


\section{Numerical experiments} \label{sec:num-exp}

\begin{figure}[htbp]
    \centering
    \includegraphics[width=0.8\textwidth]{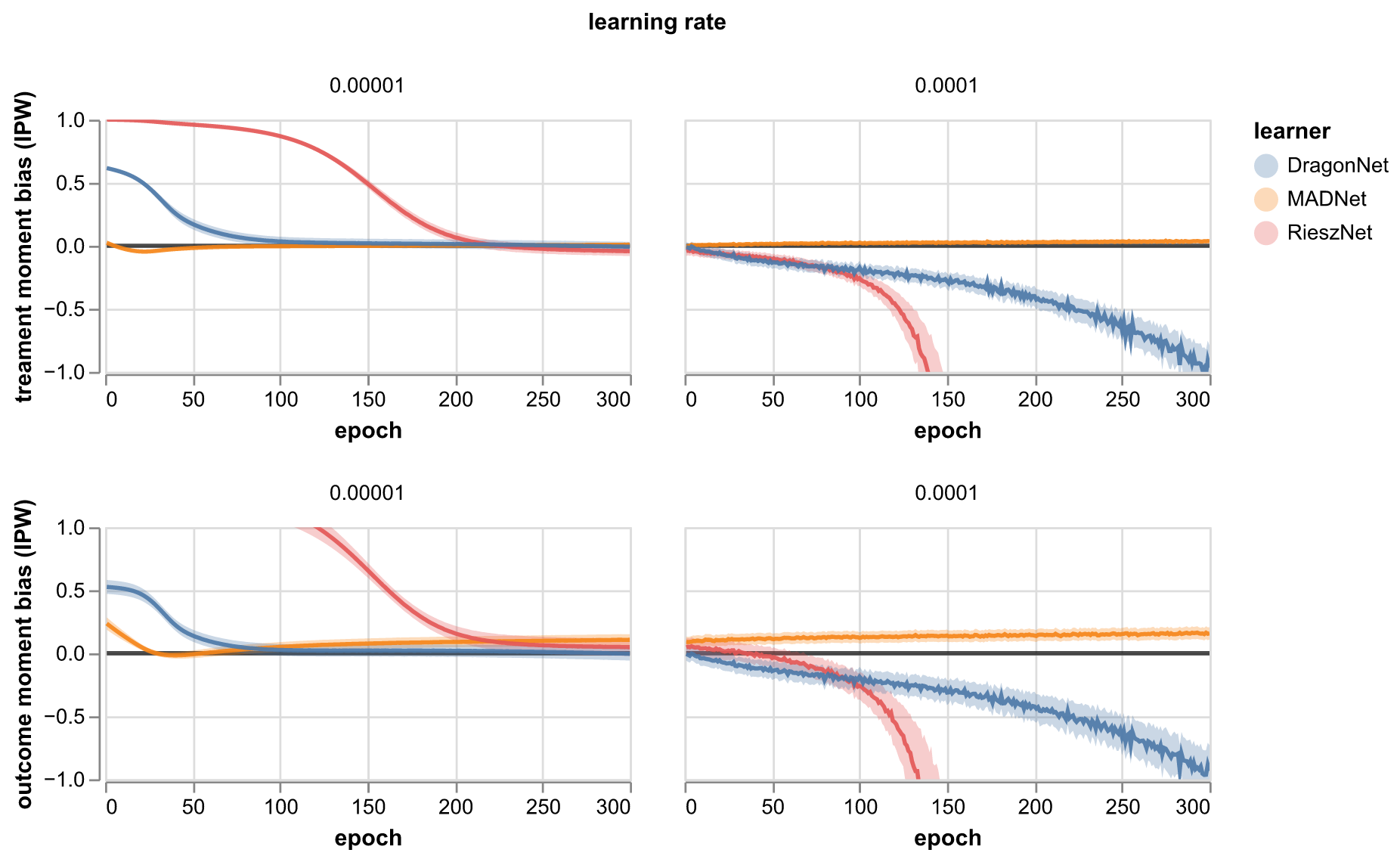}
    \caption{
    Mean and standard error of $\E_n[A\hat{\alpha}(Z)] - 1$ (top row) and of $\hat{\Psi}^{(\text{IPW})} - h_n(\mu)$ (bottom row) using 20 datasets of IHDP data where predictions are made on a 20\% validation set and the outcome is scaled by its standard deviation.
    }
    \label{fig:pred-plot}
\end{figure}


We consider ATE and ADE estimation in the following semi-synthetic data scenarios.

\textbf{IHDP:} (Infant Health and Development Program). IHDP is a randomized experiment on the effects of home visits by specialists (binary treatment, $A$) on infant cognition scores ($Y$), given 25 baseline covariates. The data consists of n = 747 infants. Synthetic outcomes are drawn from a normal distribution given $(A,X)$, as described by \citep{Hill2011}. We consider 1000 synthetic IHDP datasets in total.


\textbf{BHP:} (Blundell, Horowitz and Parey) \citep{Blundell2017}. BHP consists of 3,640 household level observations from the 2001 (U.S.) National Household Travel Survey, with the goal of estimating the price elasticity of gasoline consumption given 18 confounding variables. Price elasticity can be defined through the ADE of log price ($A$) on the log quantity of gasoline sold ($Y$). Following \citep{Chernozhukov2022}, we draw synthetic treatments from a normal distribution, with conditional mean and variance obtained from random forest predictions of the mean and variance of the true log price. Conditionally normal synthetic outcomes are then generated, given $(A,X)$, with a mean function that is cubic in treatment. The results in Table \ref{tab:ate-mae} are evaluated over 200 random seeds.


The first experimental research question is: to what extent do RR predictions from moment-constrained auto-debiasing learners satisfy oracle RR properties that are known \emph{a priori}? To answer this, we consider Inverse Probability Weighted (IPW) estimators of $\E[A\alpha(Z)] = 1$, and $\Psi$. The corresponding IPW estimators $\E_n[A\hat{\alpha}(Z)]$ and $\hat{\Psi}^{(\text{IPW})} \equiv \E_n[Y\hat{\alpha}(Z)]$ do not depend on the outcome model, thus are a convenient way of comparing RR learners. Figure \ref{fig:pred-plot} shows, using IHDP data, how the mean error evolves over gradient descent iterations (epochs) for two different learning rates. These plots show that the MADNet RR estimator converges rapidly to a stable optimum, and is therefore more robust to changes in learning rate and early stopping hyperparameters.

Next we compare absolute errors of MADNet estimators versus several alternatives: DragonNet \citep{Shi2019}, Reproducing Kernel Hilbert Space (RKHS) Embedding \citep{Singh2023}, Neural Net (NN) Embedding \citep{Xu2022}, and RieszNet \citep{Chernozhukov2022}, with only the latter applying to ADE estimation in the BHP scenario. The main proposal of our work is the Double Robust MADNet estimator, though to examine the effect of RR based bias correction, we have also provide results for the Direct (outcome model only without RR bias corrections), and IPW (RR model only) estimators.

Results in Table \ref{tab:ate-mae} show that the MADNet DR estimator has improved empirical performance (reduced mean absolute error) versus all alternatives in both scenarios. Interestingly, for the IHDP scenario, this improvement in performance is observed despite the MADNet Direct estimator performing poorly compared to its RieszNet counterpart. Moreover, the MADNet DR estimator effectively combines two estimators with poor performance (Direct and IPW) to create an estimator with good performance. One possible explanation for this could be that MADNet results in estimators for which the mixed bias term $\langle \hat{\mu} - \mu, \hat{\beta}_\perp - \beta_\perp \rangle$ is small, i.e. errors in the outcome and RR learners are uncorrelated. Further work is needed, however, to examine this effect.
Similar results are obtained when the multi-headed architectures of MADNet and RieszNet are ablated and replaced with simple feedforward NNs (Supplement Table~\ref{tab:repro-results}). Furthermore, MADNet results with a reduced constraint penalization ($\tilde{\lambda} = 1$) show slightly worse performance, highlighting the importance of satisfying the constraint in our procedure (Supplement Table~\ref{tab:hyperparam-sensitivity}).


Detailed implementation notes are provided in Supplement \ref{supp_numerical} alongside DragonNet/RieszNet results obtained using our implementation (Table~\ref{tab:repro-results}). We highlight, however, three main differences in our implementation: (i) our implementation is built on a JAX + Equinox computational stack \citep{jax2018github,kidger2021equinox}; (ii) for ADE estimation we use automatic differentiation of NN outputs, rather than a finite difference approximation; (iii) the original RieszNet uses a  complicated learning rate / early stopping scheme 
but we use a slightly simpler scheme.
Overall we found that replication via re-implementation of the RieszNet results was challenging, possibly due to the stability issues in Figure \ref{fig:pred-plot}.

\begin{table}[ht]
\centering
\caption{Absolute error (mean $\pm$ standard error) of the ATE and ADE estimates for both semi-synthetic data scenarios. 
For the RieszNet IHDP benchmark 
we report values obtained by running \texttt{RieszNet\_IHDP.ipynb} from \url{https://github.com/victor5as/RieszLearning} without modification. 
MAEs for the BHP scenario are not reported by \cite{Chernozhukov2022}, and we instead use values from our RieszNet re-implementation. 
DragonNet and RKHS/NN embedding values are retrieved from \citet[Table~1]{Xu2022}.}
\label{tab:ate-mae}
\begin{tabular}{@{}lcccc@{}}
\toprule
Estimator         & IHDP              & BHP              & Citation                 \\ \midrule
DragonNet (DR)        & 0.146 $\pm$ 0.010 & -- & \citep{Shi2019}          \\
RKHS Embedding    & 0.166 $\pm$ 0.003 & -- & \citep{Singh2023}        \\
NN Embedding      & 0.117 $\pm$ 0.002 & -- & \citep{Xu2022}           \\
RieszNet (Direct) & 0.128 $\pm$ 0.004 & 0.692 $\pm$ 0.040 & \citep{Chernozhukov2022} \\
RieszNet (IPW)    & 0.789 $\pm$ 0.036 & 0.449 $\pm$ 0.025 & \citep{Chernozhukov2022} \\
RieszNet (DR)     & 0.114 $\pm$ 0.003 & 0.428 $\pm$ 0.023 & \citep{Chernozhukov2022} \\
MADNet (Direct) & 0.504 $\pm$ 0.016 & 0.471 $\pm$ 0.026 & Proposed   \\
MADNet (IPW)    & 0.719 $\pm$ 0.039 & 0.474 $\pm$ 0.026 & Proposed   \\
MADNet (DR)     & \textbf{0.094 $\pm$ 0.002} & \textbf{0.391 $\pm$ 0.019} & Proposed   \\
\bottomrule
\end{tabular}
\end{table}

\section{Conclusion}

We present a new algorithm for estimating average moment estimands.
Our approach leverages functions for which the average moment is known \emph{a priori} to be non-zero, though the need to construct such functions may also be viewed as a limitation. Nonetheless, we contend that constructing such functions is significantly simpler than deriving the functional form of the RR, which is required for non-automatic RR learning methods. Our proposal is therefore `automatic' in the sense of not requiring complicated estimand-specific machinery. Moreover, rather than learning the full RR, as in conventional AD, we instead learn a moment-constrained function that is sufficient for debiasing the naive (direct) average moment estimator. From a practical perspective, the need to learn a constrained function requires additional techniques and hyperparameter tuning to ensure that constraints are approximately satisfied. In our work, we propose a Lagrange-type penalization method for moment-constrained learning and apply this method using multi-tasking neural networks, though experimenting with other approaches to constrained function learning represents an important direction for future study.

There are several other directions which one might extend our work. First, our set up considers estimands that represent the average moment of a regression functions, but extensions of AD to so-called generalized regressions (e.g. quantile functions) could also be considered \citep{Chernozhukov2021}. 
Second, we consider neural network learners, but similar extensions for gradient boosted trees / random forests should also be possible. 
Finally, our approach to moment-constrained function learning may be applied to other problems with stochastic constraints e.g. those related to fairness of ML predictions.



\section*{Acknowledgments}

We acknowledge insightful conversations on this topic with Alejandro Schuler and Stijn Vansteelandt.

\bibliography{refs}
\bibliographystyle{apalike} 

\newpage
\appendix

\section*{Supplement to: \mytitle}

\section{Notes on the numerical experiments}
\label{supp_numerical}

\subsection{MADNet architecture}

To ensure a fair evaluation, our proposed MADNet architecture emulates that of the RieszNet \citep{Chernozhukov2022} (see Figure \ref{fig:sharedmlps} for a schematic of the multi-headed architecture). MADNet uses a shared network of width 200 and depth 3 followed by three branches: 2 outcome networks (one per binary treatment) each of width 100 and depth 2 and another of depth zero, i.e. a linear combination of the final shared representation layer that is our $\hat{\beta}_\perp$ prediction. The constraint weight hyperparameter was set to $\Tilde{\lambda}=5$, the weight mixing parameter was set to $\rho=1$, and Exponential Linear Unit (ELU) activation functions were used throughout. Finally, outcomes $Y$ were scaled by their sample standard deviation prior to training, with predictions rescaled to the original scale using the same constant standard deviation estimate.

\textbf{Ablation study}: We compared the performance of the MADNet proposal across learner architectures, by conducting an ablation study wherein the multi-headed architecture was replaced with a fully connected MLP architecture. In particular, we used a standard feedforward network of width 200 and depth 4 along with the same hyperparameters outlined above. Results, reported in Table~\ref{tab:repro-results}, indicate that the multi-headed architecture leads to a modest reduction in mean absolute error (MAE) in all but the MADNet (IPW) estimator, and that MADNet estimators tend to outperform their RieszNet counterparts using both architectures (in terms of reduced MAE).

\textbf{Hyperparameter sensitivity}: We examine sensitivity of our proposal to the penalization strength by running the MADNet with the $\tilde{\lambda} = 1$, and other parameters unchanged. Results, reported in Table~\ref{tab:hyperparam-sensitivity} show slightly worse performance (increased MAE and increased Median absolute error), compared to results in Table~\ref{tab:repro-results}, where $\tilde{\lambda} = 5$. This suggests that a high degree of weight should be given to satisfying the moment constraint.

\subsection{MADNet training details} \label{sec:training-deets}

Numerical experiments were run on an Apple M2 Max chip with 32GB of RAM. The MADNet training procedure was also borrowed from \citet[Appendix A1]{Chernozhukov2022}, which itself was borrowed from \citet{Shi2019}. Minor modifications are outlined below. The dataset was split into a training dataset (80\%) and validation dataset (20\%), with estimation performed on the entire dataset. The training followed a two stage procedure outlined below.

\paragraph{ATE benchmarks}

\begin{enumerate}
    \item Fast training: batch size: 64, learning rate: 0.0001, maximum number of epochs: 100, optimizer: Adam, early stopping patience: 2, L2 weight decay: 0.001
    \item Fine-tuning: batch size: 64, learning rate: 0.00001, maximum number of epochs: 600, optimizer: Adam, early stopping patience: 40, L2 weight decay: 0.001
\end{enumerate}

\paragraph{ADE benchmarks}

\begin{enumerate}
    \item Fast-training: batch size: 64, learning rate: 0.001, maximum number of epochs: 100, optimizer: Adam, early stopping patience: 2, L2 weight decay: 0.001
    \item Fine-tuning: batch size: 64, learning rate: 0.0001, maximum number of epochs: 300, optimizer: Adam, early stopping patience: 20, L2 weight decay: 0.001
\end{enumerate}

The differences between the original implementations and ours are:

\begin{itemize}
    \item For ADE moment estimation, RieszNet uses a finite difference approximation to differentiate the forward pass with respect to the treatment $a$. However our implementation uses automatic differentiation provided by JAX. One of the advantages of JAX is that the ADE can be straightforwardly expressed as \texttt{jax.grad(f)(a, x)}.
    \item On top of the early stopping callback, the original RieszNet and DragonNet implementations additionally use a learning rate plateau schedule that halves the learning rate when the validation loss metric has stopped improving over a short patience of epochs (shorter than the stopping patience). Whilst we implement the same two-stage training with early stopping, we use a constant learning rate in each of the fast-training and fine-tuning phases.
    \item L2 regularization is implemented differently between RieszNet and DragonNet. DragonNet use a regularizer to apply a penalty on the layer's kernel whilst RieszNet uses an additive L2 regularization term in their loss function \citep[Equation~5]{Chernozhukov2022}. However, recent work shows that L2 regularization and weight decay regularization are not equivalent for adaptive gradient algorithms, such as Adam \citep{loshchilovDecoupledWeightDecay2019}. For this reason, we use Adam with weight decay regularization (provided by \texttt{optax.adamw}).
    \item We use a larger learning rate (0.9) for the constant additive bias parameters associated with the MLP outputs for the outcome, i.e. $f_{w,2}$ and $f_{w,3}$.
\end{itemize}

\begin{table}[hbt!]
    \centering
    \caption{Full reproduction results for our own implementation of each learner/estimator. Here + SRR, refers to estimator which use the outcome model $\tilde{g}$ described in \cite{Chernozhukov2022}.}
    \resizebox{\textwidth}{!}{%
    \begin{tabular}{lllrrr}
    \toprule
     &  &  & Mean Absolute Error (MAE) & Median Absolute Error & Standard Error in MAE \\
    Dataset & Estimator & Architecture &  &  &  \\
    \midrule
    \multirow[t]{18}{*}{BHP} & \multirow[t]{2}{*}{MADNet (DR)} & Fully connected & 0.417 & 0.394 & 0.021 \\
     &  & Multiheaded & 0.391 & 0.346 & 0.019 \\
    \cmidrule(lr){2-6}
     & \multirow[t]{2}{*}{MADNet (Direct)} & Fully connected & 0.512 & 0.427 & 0.029 \\
     &  & Multiheaded & 0.471 & 0.424 & 0.026 \\
    \cmidrule(lr){2-6}
     & \multirow[t]{2}{*}{MADNet (IPW)} & Fully connected & 0.407 & 0.352 & 0.023 \\
     &  & Multiheaded & 0.474 & 0.404 & 0.026 \\
    \cmidrule(lr){2-6}
     & \multirow[t]{2}{*}{RieszNet (DR + SRR)} & Fully connected & 0.447 & 0.370 & 0.024 \\
     &  & Multiheaded & 0.428 & 0.355 & 0.023 \\
    \cmidrule(lr){2-6}
     & \multirow[t]{2}{*}{RieszNet (DR)} & Fully connected & 0.447 & 0.372 & 0.024 \\
     &  & Multiheaded & 0.428 & 0.353 & 0.023 \\
    \cmidrule(lr){2-6}
     & \multirow[t]{2}{*}{RieszNet (Direct + SRR)} & Fully connected & 0.771 & 0.637 & 0.041 \\
     &  & Multiheaded & 0.724 & 0.617 & 0.042 \\
    \cmidrule(lr){2-6}
     & \multirow[t]{2}{*}{RieszNet (Direct)} & Fully connected & 0.733 & 0.619 & 0.039 \\
     &  & Multiheaded & 0.692 & 0.585 & 0.040 \\
    \cmidrule(lr){2-6}
     & \multirow[t]{2}{*}{RieszNet (IPW + SRR)} & Fully connected & 0.477 & 0.432 & 0.025 \\
     &  & Multiheaded & 0.449 & 0.384 & 0.025 \\
    \cmidrule(lr){2-6}
     & \multirow[t]{2}{*}{RieszNet (IPW)} & Fully connected & 0.477 & 0.432 & 0.025 \\
     &  & Multiheaded & 0.449 & 0.384 & 0.025 \\
    \cmidrule(lr){1-6}
    \multirow[t]{24}{*}{IHDP} & DragonNet (DR + SRR) & Multiheaded & 0.101 & 0.085 & 0.003 \\
    \cmidrule(lr){2-6}
     & DragonNet (DR) & Multiheaded & 0.100 & 0.084 & 0.002 \\
    \cmidrule(lr){2-6}
     & DragonNet (Direct + SRR) & Multiheaded & 0.124 & 0.098 & 0.004 \\
    \cmidrule(lr){2-6}
     & DragonNet (Direct) & Multiheaded & 0.123 & 0.098 & 0.004 \\
    \cmidrule(lr){2-6}
     & DragonNet (IPW + SRR) & Multiheaded & 0.262 & 0.233 & 0.006 \\
    \cmidrule(lr){2-6}
     & DragonNet (IPW) & Multiheaded & 0.262 & 0.233 & 0.006 \\
    \cmidrule(lr){2-6}
     & \multirow[t]{2}{*}{MADNet (DR)} & Fully connected & 0.096 & 0.079 & 0.003 \\
     &  & Multiheaded & 0.094 & 0.076 & 0.002 \\
    \cmidrule(lr){2-6}
     & \multirow[t]{2}{*}{MADNet (Direct)} & Fully connected & 0.527 & 0.383 & 0.018 \\
     &  & Multiheaded & 0.504 & 0.367 & 0.016 \\
    \cmidrule(lr){2-6}
     & \multirow[t]{2}{*}{MADNet (IPW)} & Fully connected & 0.680 & 0.263 & 0.037 \\
     &  & Multiheaded & 0.719 & 0.277 & 0.039 \\
    \cmidrule(lr){2-6}
     & \multirow[t]{2}{*}{RieszNet (DR + SRR)} & Fully connected & 0.119 & 0.091 & 0.004 \\
     &  & Multiheaded & 0.109 & 0.088 & 0.003 \\
    \cmidrule(lr){2-6}
     & \multirow[t]{2}{*}{RieszNet (DR)} & Fully connected & 0.119 & 0.090 & 0.004 \\
     &  & Multiheaded & 0.109 & 0.089 & 0.003 \\
    \cmidrule(lr){2-6}
     & \multirow[t]{2}{*}{RieszNet (Direct + SRR)} & Fully connected & 0.135 & 0.099 & 0.006 \\
     &  & Multiheaded & 0.126 & 0.102 & 0.004 \\
    \cmidrule(lr){2-6}
     & \multirow[t]{2}{*}{RieszNet (Direct)} & Fully connected & 0.135 & 0.105 & 0.004 \\
     &  & Multiheaded & 0.118 & 0.099 & 0.003 \\
    \cmidrule(lr){2-6}
     & \multirow[t]{2}{*}{RieszNet (IPW + SRR)} & Fully connected & 0.690 & 0.304 & 0.035 \\
     &  & Multiheaded & 0.665 & 0.300 & 0.036 \\
    \cmidrule(lr){2-6}
     & \multirow[t]{2}{*}{RieszNet (IPW)} & Fully connected & 0.690 & 0.304 & 0.035 \\
     &  & Multiheaded & 0.665 & 0.300 & 0.036 \\
    \bottomrule
    \end{tabular}
    }
    \label{tab:repro-results}
\end{table}

\begin{table}[hbt!]
    \centering
    \caption{Numerical experiment results for the Multiheaded MADNet procedure with $\tilde{\lambda}=1$.}
    \resizebox{\textwidth}{!}{%
    \begin{tabular}{llrrr}
    \toprule
     &  & Mean Absolute Error (MAE) & Median Absolute Error & Standard Error in MAE \\
    Dataset & Estimator &  &  &  \\
    \midrule
    \multirow[t]{3}{*}{BHP} & MADNet (DR) & 0.407 & 0.370 & 0.021 \\
     & MADNet (Direct) & 0.479 & 0.415 & 0.025 \\
     & MADNet (IPW) & 0.544 & 0.459 & 0.031 \\
    \cmidrule(lr){1-5}
    \multirow[t]{3}{*}{IHDP} & MADNet (DR) & 0.098 & 0.077 & 0.003 \\
     & MADNet (Direct) & 0.519 & 0.382 & 0.016 \\
     & MADNet (IPW) & 0.712 & 0.283 & 0.038 \\
    \bottomrule
    \end{tabular}
    }
    \label{tab:hyperparam-sensitivity}
\end{table}

\subsection{Naive Lagrangian optimization}

\begin{figure}[htbp]
    \centering
    \includegraphics[width=\textwidth]{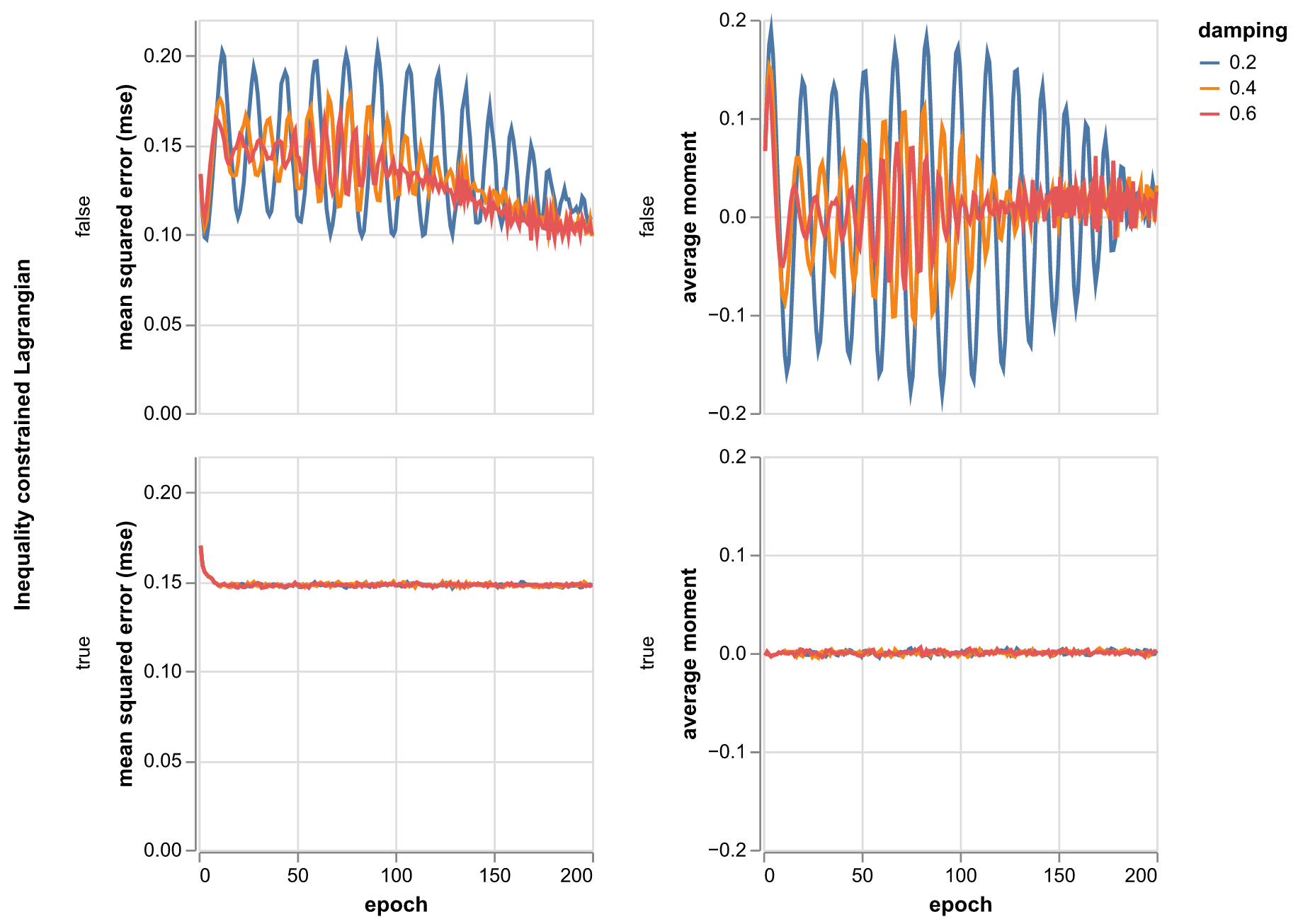}
    \caption{
    Top row: Low damping coefficients in the basic differential multiplier method (BDMM) \citep{Platt1987} lead to oscillatory behavior around the saddle point solution when the optimisation problem is formulated as an equality constrained Lagrangian. Bottom row: Using the inequality constrained Lagrangian approach described in the main paper results in more stable training and constraint satisfaction. A single dataset from the IHDP data is used to showcase this behavior over 200 epochs.
    }
    \label{fig:damping-plot}
\end{figure}

We consider the basic differential multiplier method (BDMM), as described by Platt et al. \citep{Platt1987}. The authors introduce a so-called damping term $\delta \geq 0$ to the Lagrangian in \eqref{pop_lagrangian} to obtain the Lagrangian
\begin{align*}
    \Lagrangian_\delta(f, \lambda) \equiv \E\left[\{\beta(Z) - f(Z)\}^2\right] + \lambda h(f) + \delta h^2(f),
\end{align*}
with \eqref{pop_lagrangian} recovered by setting $\delta = 0$. Note that when the moment constraint is satisfied, i.e.$h(f) = 0$, then $\Lagrangian_\delta$ does not depend on $\delta$. In Figure \ref{fig:damping-plot} we see how Naively performing gradient ascent on $\lambda$ and gradient descent over $f$ results in oscillatory behavior. Similar behavior is also observed in the literature on adversarial learning, see e.g. \citep{Schafer2019, Mokhtari2020}.

\section{Short notes and proofs}

\subsection{First-order remainder under the standard theory}
\label{proof:first-order}
Claim: $\E[\hat{\varphi}(W) - \Psi] = -\langle \hat{\mu} - \mu, \hat{\alpha} - \alpha \rangle$ where $\hat{\varphi}(W) = m(\hat{\mu}, W) + \hat{\alpha}(Z)\{Y - \hat{\mu}(Z)\}$.

Proof:
\begin{align*}
    & \E[ m(\hat{\mu}, W) + \hat{\alpha}(Z)\{Y - \hat{\mu}(Z)\}  - \Psi ] \\
    = & \E[ m(\hat{\mu}, W) + \hat{\alpha}(Z)\{\mu(Z) - \hat{\mu}(Z)\}  - m(\mu, W) ] \\
    = & \E[ m(\hat{\mu} - \mu, W) - \hat{\alpha}(Z)\{ \hat{\mu}(Z) - \mu(Z)\} ] \\
    = & \langle \hat{\mu} - \mu, \alpha \rangle - \langle \hat{\mu} - \mu, \hat{\alpha} \rangle  \\
    = & -\langle \hat{\mu} - \mu, \hat{\alpha} - \alpha \rangle.
\end{align*}

\subsection{Second-order remainder under the standard theory}
Claim: If $\hat{\mu}$ and $\hat{\alpha}$ are consistent estimators for $\mu$ and $\alpha$ obtained from an independent sample, and there exists a constant $M$ such that $\alpha^2(Z) < M$ and $\Var(Y|Z) < M$ almost surely, then $G_n[\hat{\varphi}(W) - \varphi(W)] = o_p(1)$.

Proof:
\begin{align*}
    G_n[\hat{\varphi}(W) - \varphi(W)]=
    &+ G_n[m(\hat{\mu} - \mu, W)] \\
    &- G_n\left[\{\hat{\alpha}(Z) - \alpha(Z)\}\{\hat{\mu}(Z) - \mu(Z)\}\right] \\
    &- G_n\left[\alpha(Z)\{\hat{\mu}(Z) - \mu(Z)\}\right] \\
    &+ G_n\left[\{\hat{\alpha}(Z) - \alpha(Z)\}\{Y - \mu(Z)\}\right] 
\end{align*}
By the central limit theorem, these empirical processes are $o_p(1)$ when the following expressions are $o_p(1)$
\begin{align*}
    &\E\left[m^2(\hat{\mu} - \mu, W)\right]\\
    &\E\left[\{\hat{\alpha}(Z) - \alpha(Z)\}^2\{\hat{\mu}(Z) - \mu(Z)\}^2\right] \\
    &\E\left[\alpha^2(Z)\{\hat{\mu}(Z) - \mu(Z)\}^2\right]\\
    &\E\left[\{\hat{\alpha}(Z) - \alpha(Z)\}^2\{Y - \mu(Z)\}^2\right]
\end{align*}
The first two terms are $o_p(1)$ by consistency of $\hat{\alpha}$ and $\hat{\mu}$, for the final two terms
\begin{align*}
    \E\left[\alpha^2(Z)\{\hat{\mu}(Z) - \mu(Z)\}^2\right] &< M ||\hat{\mu} - \mu||^2 \\
    \E\left[\{\hat{\alpha}(Z) - \alpha(Z)\}^2\Var(Y|Z)\right] &< M ||\hat{\alpha} - \alpha||^2
\end{align*}
hence, these are also $o_p(1)$ by consistency.

Remark: The requirement for estimator independence can be relaxed if one makes Donsker class assumptions instead.

\subsection{Proof of Theorem \ref{main_theorem}}
\label{main_theorem_proof}

Consider \eqref{von_miesz} with $\hat{\psi} = h_n(\hat{\mu}^*)$ and $\hat{\varphi}(W) = m(\hat{\mu}^*, W) + k\{\beta(Z) - \hat{\beta}_\perp(Z)\}\{Y - \hat{\mu}^*(Z)\}$, where we use the shorthand
\begin{align*}
    k = \frac{h(\beta)}{||\beta - \beta_\perp||^2}
\end{align*}
so that, by \eqref{new_ident_1}, $\alpha(Z) = k\{\beta(Z) - \beta_\perp(Z)\}$ and $\varphi(W) = m(\mu, W) + k\{\beta(Z) - \beta_\perp(Z)\}\{Y - \mu(Z)\}$.
Under this parameterization, the plug-in bias on the right hand side of \eqref{von_miesz} is
\begin{align*}
    \sqrt{n} \mathbb{E}_n[\hat{\varphi}(W) - \hat{\psi}]= \sqrt{n} k \mathbb{E}_n[\{\beta(Z) - \hat{\beta}_\perp(Z)\}\{Y - \hat{\mu}^*(Z)\}] = 0
\end{align*}
Applying the result in Supplement \ref{proof:first-order}, the first-order remainder on the right hand side of \eqref{von_miesz} is
\begin{align*}
    \sqrt{n}\E[\hat{\varphi}(W) - \Psi] = \sqrt{n} k \langle \hat{\mu}^* - \mu, \hat{\beta}_\perp - \beta_\perp \rangle = o_p(1)
\end{align*}
Finally the second-order remainder on the right hand side of \eqref{von_miesz} is
\begin{align*}
    G_n \left[\hat{\varphi}(W) - \varphi(W)\right] &= G_n[m(\hat{\mu}^* - \mu, W)] \\
    &+ k G_n\left[\{\hat{\beta}_\perp(Z) - \beta_\perp(Z)\}\{\hat{\mu}^*(Z) - \mu(Z)\}\right] \\
    &- k G_n\left[\{\beta(Z) - \beta_\perp(Z)\}\{\hat{\mu}^*(Z) - \mu(Z)\}\right] \\
    &-k G_n\left[\{\hat{\beta}_\perp(Z) - \beta_\perp(Z)\}\{Y - \mu(Z)\}\right] 
\end{align*}
which is $o_p(1)$.

We have shown that plug-in bias, first-order remainder and second-order remainder in \eqref{von_miesz} are each $o_p(1)$, hence $h_n(\hat{\mu}^*)$ is RAL.

\subsection{Sufficiency of learning conditional on the unscaled RR}
\label{proof_of_suff}
Claim: $\Psi = h(\eta)$, where
\begin{align*}
    \eta(z) \equiv \E[Y|\beta(Z) - \beta_\perp(Z) = \beta(z) - \beta_\perp(z)].
\end{align*}
Proof:
\begin{align*}
    \Psi &= \E[Y \alpha(Z)] \\
    &= \frac{h(\beta) \E[Y \{\beta(Z) - \beta_\perp(Z)\}]}{||\beta - \beta_\perp||} \\
    &= \frac{h(\beta) \E[\eta(Z) \{\beta(Z) - \beta_\perp(Z)\}]}{||\beta - \beta_\perp||} \\
    &= \E[\eta(Z) \alpha(Z)]
\end{align*}
where in the third step we apply the law of iterated expectation.

\subsection{Proof of orthogonality representation}
\label{proof_of_ortho}

Claim: Letting $\mu_\perp = \argmin_{f\in \mathcal{C}^{\perp}} ||\mu - f||$
\begin{align*}
    \mu(z) = \mu_{\perp}(z) + \frac{\Psi}{h(\beta)} \{\beta(z) - \beta_{\perp}(z)\}.
\end{align*}

Proof:
Note that $\mathcal{C}^\perp = \{f\in \mathcal{H} \mid  \langle f, \alpha \rangle = 0 \}$ then by Hilbert's projection theorem, $\mu_\perp$ exists, with
\begin{align*}
    \mu_\perp(z) \equiv \mu(z) - \frac{\langle \mu, \alpha \rangle}{||\alpha||^2} \alpha(z) 
    \quad\iff\quad \mu(z) = \mu_\perp(z) + \frac{\Psi}{h(\alpha)} \alpha(z) 
\end{align*}
Where we use $\langle \mu, \alpha \rangle = \Psi$ and $||\alpha||^2 = h(\alpha)$. Applying \eqref{new_ident_1} completes the proof.

\end{document}